\documentclass{article}
\usepackage{spconf,amsmath,graphicx,amssymb}
\usepackage{multirow}
\usepackage[table]{xcolor}
\usepackage{tabularx, etoolbox}

\usepackage[pagebackref,breaklinks,colorlinks]{hyperref}
\newcolumntype{P}[1]{>{\centering\arraybackslash}p{#1}}
\usepackage[capitalize]{cleveref}
\crefname{section}{Sec.}{Secs.}
\Crefname{section}{Section}{Sections}
\Crefname{table}{Table}{Tables}
\crefname{table}{Tab.}{Tabs.}

\title{TSANet: Temporal and Scale Alignment \\for Unsupervised Video Object Segmentation }
\name{Seunghoon Lee$^1$\quad Suhwan Cho$^1$\quad Dogyoon Lee$^1$\quad Minhyeok Lee$^1$\quad Sangyoun Lee$^{1,2}$}
\address{$^1$Yonsei University, Korea \\ $^2$Korea Institute of Science and Technology (KIST)}
\vspace{-1.0cm}
\begin{document}
%
\maketitle
\begin{abstract}
Unsupervised Video Object Segmentation (UVOS) refers to the challenging task of segmenting the prominent object in videos without manual guidance. In other words, the network detects the accurate region of the target object in a sequence of RGB frames without prior knowledge. In recent works, two approaches for UVOS have been discussed that can be divided into: appearance and appearance-motion based methods. Appearance based methods utilize the correlation information of inter-frames to capture target object that commonly appears in a sequence. However, these methods does not consider the motion of target object due to exploit the correlation information between randomly paired frames. Appearance-motion based methods, on the other hand, fuse the appearance features from RGB frames with the motion features from optical flow. Motion cue provides useful information since salient objects typically show distinctive motion in a sequence. However, these approaches have the limitation that the dependency on optical flow is dominant. In this paper, we propose a novel framework for UVOS that can address aforementioned limitations of two approaches in terms of both time and scale. Temporal Alignment Fusion aligns the saliency information of adjacent frames with the target frame to leverage the information of adjacent frames. Scale Alignment Decoder predicts the target object mask precisely by aggregating differently scaled feature maps via continuous mapping with implicit neural representation. We present experimental results on public benchmark datasets, DAVIS 2016 and FBMS, which demonstrate the effectiveness of our method. Furthermore, we outperform the state-of-the-art methods on DAVIS 2016.
\end{abstract}
\vspace{-0.1cm}
\vspace{-0.4cm}
\section{Introduction}
\label{sec:intro}

Video Object Segmentation (VOS) is a vital task in the field of computer vision. VOS aims to segment accurate regions of the target object by pixel unit. It has been widely applied to various fields such as surveillance, video editing, medicine, and autonomous driving. More specifically, Unsupervised VOS (UVOS) segments the most salient object without any prior knowledge at the test time. This is a challenging task in that the network should recognize a target object, which is either the most prominent moving object or has the most consistent appearance in a sequence without using a manual guidance.

\begin{figure*}[t!]
    \centerline{\includegraphics[width=\textwidth,height=6.42cm]{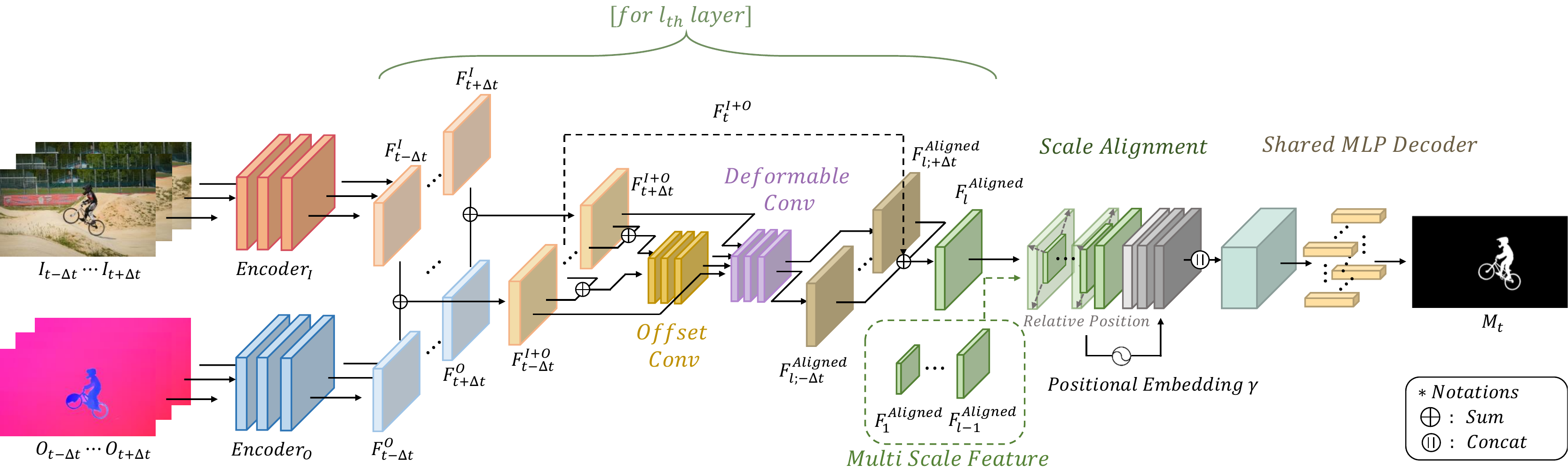}}
    \vspace{-0.3cm}
    \caption{Overview of TSANet architecture.}
    \label{fig:figure_1}
    \vspace{-0.5cm}
\end{figure*}

In recent years, several studies based on deep learning have been progressed into two streams for addressing UVOS: appearance-base methods and appearance-motion based methods. Appearance-based methods~\cite{cosnet,cfam,f2net}, use only appearance features extracted from multiple frames. These methods proposed the module that captures the saliency information based on attention mechanism between target frame and other frames. These approaches can fully use contextual information including semantic details of inter-frames, regardless of frame order. However, these approaches focus on the correlation between saliency features from different frames without considering the motion of the target object. Furthermore, these methods tend to vary in performance depending on which frame is used as a reference at test time. Appearance-motion based methods~\cite{matnet,rtnet,dtt,amc,tmo} use the target frame as well as corresponding optical flow (called "motion cue"), which is calculated using optical flow networks~\cite{flownet,raft}. Motion cue enables the model to capture the prominent region of the target object. By combining appearance features with motion features, it is possible to acquire distinct region of salient object whose motion is prominent. However, these methods have an intrinsic drawback that the dependency on optical flow is high. If the target object shows significant motion blur or occlusion, the optical flow quality is inferior, which makes it difficult to determine the target object accurately. This limitation has been addressed with several methods~\cite{amc,rtnet}. \cite{amc} exploits multi modality fusion module that, adaptively, fuses appearance with motion features by scoring each feature. \cite{rtnet} utilizes bidirectional optical flow to fully take advantage of the motion cues between two frames. However, one problem still remain of these methods: model cannot use the contextual information of consecutive RGB frames considering the characteristics of the video sequence.

To tackle the aforementioned problems, we propose a novel framework, Temporal and Scale Alignment Network (TSANet), which can utilize both contextual and motion information from adjacent frames. Our framework consists of two modules, Temporal Alignment Fusion~(TAF) and Scale Alignment Decoder~(SAD). TAF aligns the information of adjacent frames to the target frame to propagate contextual information in multi-layer features extracted from each residual block of ResNet~\cite{resnet}. Then, SAD aggregates the features with different sizes which are aligned from TAF. In particular, we introduce scale alignment method, which is based on implicit neural representation. We generate a continuous feature map by embedding the relative coordinates and feature values to interpret the correlation information between relative coordinates when upscaling the aligned features. We demonstrate the effectiveness of our method on benchmark datasets, DAVIS 2016 and FBMS. Our proposed framework show the state-of-the-art performance in DAVIS 2016 dataset and compromising results in FBMS dataset.\\Our contributions are as follows:
\vspace{-0.2cm}

\begin{itemize}
\item We propose Temporal Alignment Fusion (TAF) that aligns features of adjacent frames to target features for leveraging the contextual information of adjacent frames to the target frame.
\item To aggregate features which have different scales, we propose Scale Alignment Decoder (SAD), introducing implicit neural representation using relative coordinate information that occurs in alignment process.
\item Our framework (TSANet) shows the state-of-the performance on DAVIS 2016 with 87.4\% $\mathcal{J}\&\mathcal{F}\;$Mean score.
\end{itemize}
\vspace{-0.6cm}
\section{Proposed Method}
\label{sec:method}
\vspace{-0.2cm}
\subsection{Overview of the Network Architecture}
The overview of TSANet is shown in \figurename~\ref{fig:figure_1}. Given an input video sequence of length N, we aim to segment the target objects in each frame $\{I_1,...,I_N\}$, where $t \in \{1,\cdots,N\}$. RGB frames are denoted as $I_t \in \mathbb{R}^{H \times W \times 3}$ and corresponding optical flow maps (converted into 3 channels) are denoted as
$O_t \in \mathbb{R}^{H \times W \times 3}$. For predicting object at $I_{t}$, we use adjacent two RGB frames and optical flow maps, $I_{t-1},I_{t+1}, O_{t-1},O_{t+1}$.

We employ two branches of the encoders, $E_I$ and $E_O$
, in parallel scheme to extract appearance and motion features, $F^I$ and $F^O$, respectively. In order to aggregate useful information from frames adjacent to the target frame, we implement TAF on the consolidated features $F^{I+O}$ by combining the appearance and motion features in each $l_{th}$ layer. The combined features from the adjacent frames $F^{I+O}_{t+\Delta t}$ are aligned to combined features from target frame $F^{I+O}_{t}$ via TAF. Finally, SAD interprets the continuous feature maps introducing implicit neural representation for resolution-free manner using positional embedding of relative coordinates between the features with different scales.
\vspace{-0.45cm}

\begin{table*}[t!]
\small
\centering
{
\begin{tabular}{P{3cm}|P{2cm}|P{0.6cm}|P{1.1cm}P{1.1cm}P{1.1cm}|P{1.1cm}P{1.1cm}P{1.1cm}|P{1.1cm}}
\hline
\multirow{2}{*}{Model} & \multirow{2}{*}{Publication} & \multirow{2}{*}{PP} & \multicolumn{3}{c|}{$\mathcal{J}$}   & \multicolumn{3}{c|}{$\mathcal{F}$}   & $\mathcal{J\&F}$   \\ 
\cline{4-10} 
&                              &                           & Mean↑ & Recall↑ & Decay↓ & Mean↑ & Recall↑ & Decay↓ & Mean↑ \\ \hline
AGS~\cite{ags}                    & CVPR'19                      & $\checkmark$                  & 79.7  & 91.1    & 1.9    & 77.4  & 85.8    & 1.6    & 78.6  \\
COSNet~\cite{cosnet}                 & CVPR'19                      & $\checkmark$                 & 80.5  & 93.1    & 4.4    & 79.5  & 89.5    & 5      & 80.0  \\
ADNet~\cite{adnet}                  & ICCV'19                      & $\checkmark$                   & 81.7  & -    & -    & 80.5  & -    & -      &  81.1 \\
AGNN~\cite{agnn}                  & ICCV'19                      & $\checkmark$                   & 80.7  & 94.0    & 0.0    & 79.1  & 90.5    & 0.0      & 79.9  \\
MATNet~\cite{matnet}                  & AAAI'20                      & $\checkmark$                  & 82.4  & 94.5    & 3.8    & 80.7  & 90.2    & 4.5      & 81.5  \\
DFNet~\cite{dfnet}                  & ECCV'20                      & $\checkmark$                   & 83.4  & 94.4    & 4.2    & 81.8  & 89.0    & 3.7      & 82.6  \\
3DC-Seg~\cite{3dc}                  & BMVC'20                      & $\checkmark$                  & 84.2  & 95.8    & 7.4    & 84.3  & 92.4    & 5.5      & 84.2  \\
F2Net~\cite{f2net}                  & AAAI'21                      &                    & 83.1  & 95.7    & 0.0    & 84.4  & 92.3    & 0.8      & 83.7  \\
RTNet~\cite{rtnet}                  & CVPR'21                      & $\checkmark$                   & \cellcolor{yellow!25}85.6  & 96.1    & -    & 84.7  & 93.8    & -      & 85.2  \\
FSNet~\cite{fsnet}                  & ICCV'21                      & $\checkmark$                  & 83.4  & 94.5    & 3.2    & 83.1  & 90.2    & 2.6      & 83.3  \\
TransportNet~\cite{dtt}                 & ICCV'21                      &                  & 84.5  &  -  & -    & 85.0  & -    & -      & 84.8  \\
AMC-Net~\cite{amc}                  & ICCV'21                      & $\checkmark$                   & 84.5  & 96.4    & 2.8    & 84.6  & 93.8    & 2.5      & 84.6  \\
CFAM~\cite{cfam}                 & WACV'22                      &                   & 83.5  & -    & -    & 82.0  & -    & -      & 82.8  \\
D$^2$Conv3D~\cite{d2}                 & WACV'22          &                              & 85.5  & -    & -    & 86.5  & -    & -      & 86.0  \\
IMP~\cite{imp}                 & AAAI'22          &                              & 84.5  & 92.7    & 2.8    & \cellcolor{yellow!25}86.7  & 93.3    & 0.8      &   85.6\\
PMN~\cite{pmn}                  & WACV'23                      &      $\checkmark$             & 85.4  & -    & -    & 86.4  & -    & -      & 85.9  \\
TMO~\cite{tmo}                  & WACV'23                      &                   & \cellcolor{yellow!25}85.6  & -    & -    & 86.6  & -    & -      & \cellcolor{yellow!25}86.1  \\ \hline
\textbf{Ours}                  &                       &                    & \cellcolor{orange!25}\textbf{86.6}  & 95.7    & 0.0    & \cellcolor{orange!25}\textbf{88.3}  & 94.3    & 0.0      & \cellcolor{orange!25}\textbf{87.4}  \\ 
\hline
\end{tabular}
}
\vspace{-0.3cm}
\caption{Quantitative results on DAVIS 2016. PP indicates post-processing. Each color denotes \colorbox{orange!25}{\textbf{best}} and \colorbox{yellow!25}{second} results.}
\label{tab:quantitative}
\vspace{-0.4cm}
\end{table*}

\subsection{Temporal Alignment Fusion}
Inspired by \cite{dim}, we propose Temporal Alignment Fusion (TAF) to capture the information of the target object using the information from consecutive frames. We align the adjacent frames to the target frame via deformable convolution~\cite{deform}. TAF is defined as:
\vspace{-0.7cm}

\begin{equation}
\mathcal{F}^{Aligned}_{t+\Delta t\rightarrow t}(p)=\sum^{k-1}_{i=0} \left(w_i\cdot m_i\right) \times \mathcal{F}^{I+O}_{t+\Delta t}\left(p+\Delta p_i\right),
\end{equation}
\vspace{-0.2cm}

\noindent where $p$ and $w_i$ indicate a certain position of feature maps and kernel weights of deformable convolution filters whose kernel sizes are $\sqrt{k}\times \sqrt{k}$, respectively. Offset $\Delta p_i$ and offset mask $m_i$ are acquired from convolution on the concatenated features of $F^{I+O}_t$ and $F^{I+O}_{t+\Delta t}$ as follows:

\begin{equation}
 \Delta p_i, m_i = \operatorname{Conv}\left(F^{I+O}_{t+\Delta t} \oplus F^{I+O}_{t}\right),
\end{equation}
\noindent where $\Delta t \in \left \{ -1,1 \right \}$. $\Delta p_i$ indicates where the position of adjacent frames is aligned to the target frame to aggregate local information of adjacent frames. In each scale of features extracted from the encoder, we implement TAF to obtain $N_l$ aligned features which have different scales, where $N_l$ and $l$ denote the number of layers and $l_{th}$ layer, respectively. Thus the aligned features are formulated as follows:
\vspace{-0.5cm}

\begin{equation}
F^{Aligned}_l = \operatorname{Conv}\left(\sum_{\Delta t}\mathcal{F}^{Aligned}_{l;t+\Delta t\rightarrow t}(p)\right),\:l \in \left \{1,\cdot \cdot ,4  \right \}.
\end{equation}

We exploit TAF for features of each layer from high-level features which contain semantic information to low-level features which have detailed texture information of the scene. $N_l$ aligned features are forwarded to SAD. 
\vspace{-0.3cm}

\subsection{Scale Alignment Decoder}
To handle the multi-scale feature, we propose Scale Alignment Decoder (SAD) inspired by recent publications~\cite{ifa, vinr}. In these studies, implicit neural representation shows promising results on many computer vision tasks, i.e., video frame interpolation, semantic segmentation, super-resolution. $\{F^{align}_l\}_{l=1}^{4}$ denotes aligned features with different scales. Then we define the continuous feature map ${F_c}$ which is aggregated features from different scales through MLP layers. To assign the position of aligned features to the continuous feature map $F_c$, we normalize the coordinate $x$ at $F_c$ and all $F^{align}_l$. We then match the position $x_c$ of $F_c$ with the position $x_a$of $F^{align}_l$ using nearest neighbor based on euclidean distance. Relative coordinate values are obtained from this process which match the position $x_a$ to $x_c$. This process can be formulated as :  
\vspace{-0.7cm}

\begin{equation}
F_c\left(x_c\right)=MLP_\theta \left(Cat\{z^l_a~||~\gamma (x_c-x^l_a)\}^4_{l=1}\right),
\end{equation}
\noindent
where ${MLP}_\theta$ denotes parameterized decoding layers. $Cat$ and $||$ both denote concatenation for clarity. ${z^{l}_a}$ is a feature value at the position $x^{l}_a$ and $\gamma$ denotes positional embedding function on relative coordinate. Therefore, each pixel of $F_c$ contains information about features of all scales and their relative coordinates. The continuous feature map, which takes into account relative coordinate and corresponding feature vectors, are decoded to predict whether each pixel is a salient object or not at per-pixel level through MLP. 

\begin{figure*}[t!]
    \centerline{\includegraphics[width=\textwidth,height=7cm]{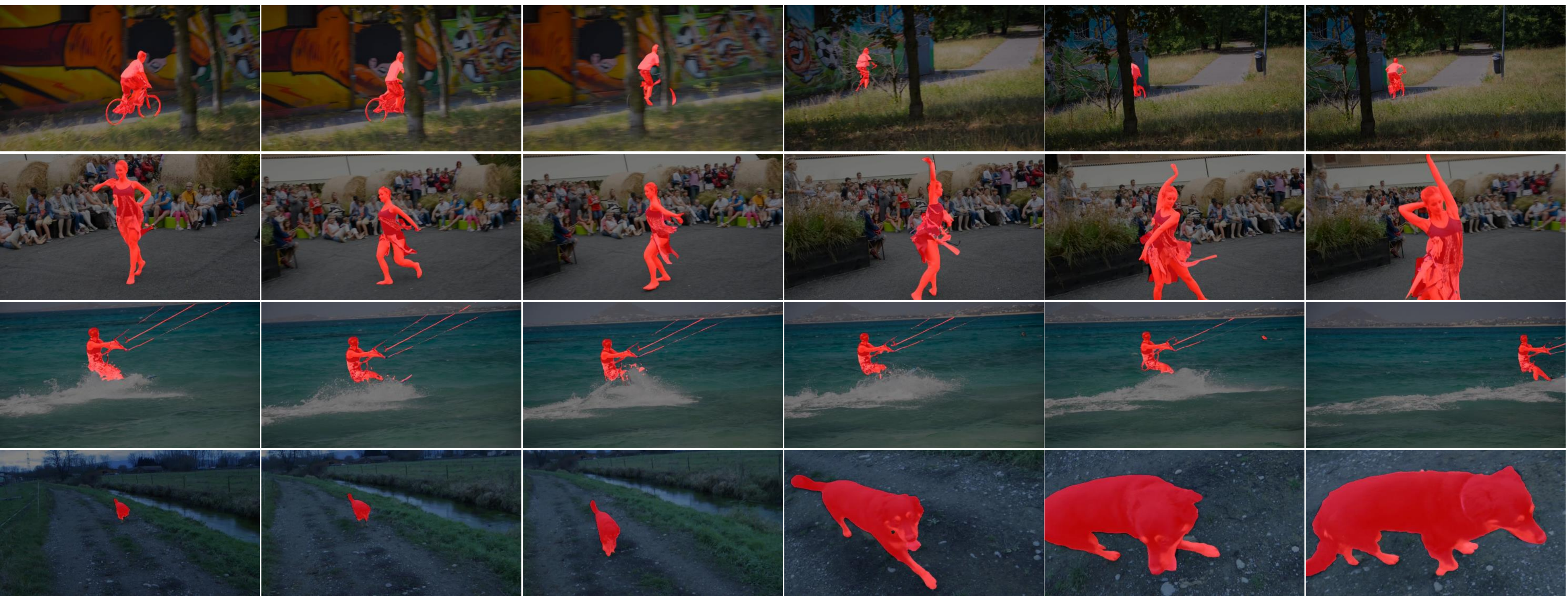}}
    \vspace{-0.4cm}
    \caption{Qualitative results. From top to bottom: bmx-trees, dance-twirl and kite-surf from DAVIS 2016, dogs02 from FBMS}
    \label{fig:qualitative} 
    \vspace{-0.3cm}
\end{figure*}

\vspace{-0.3cm}
\section{Experiments}
\label{sec:exp}
\subsection{Experimental Setup}

\textbf{DAVIS 2016}~\cite{davis} consists of 50 high-quality videos of $480p \times 720p$. There are 30 sequences for training and 20 sequences for validation. Each frame contains densely-annotated ground truth of foreground objects. Following prior works~\cite{matnet,rtnet,tmo,dtt}, we evaluate our method on two metrics, $\mathcal{J}\:$and$\: \mathcal{F}$ called as region similarity,  boundary accuracy, respectively. \\
\noindent
\textbf{FBMS}~\cite{fbms} includes 59 videos, with sparsely annotated ground truth. There are 29 training videos and 30 test videos. We evaluate our method on FBMS with respect to region similarity $\mathcal{J}$.
\vspace{-0.4cm}

\subsection{Implementation Details}
We compute the optical flow of all video sequences using RAFT \cite{raft}. Following TMO~\cite{tmo}, we adopt joint training strategy which uses saliency datasets DUTS~\cite{duts} and training sequences of DAVIS 2016. We use ResNet101 as backbone. We use multi-scale training using 4 different sizes as $\{384, 448, 480, 512\}$ and batch size is set to 4. We apply augmentation to make a single static RGB images of DUTS into 3 frames like video following \cite{otvm}. During joint training, two datasets are changed every 128 iterations. We optimize our network using cross-entropy loss and use Adam as the optimizer. We fix the learning rate as $1e^-5$. Training is implemented on 2 RTX 3090 GPUs.\\
\vspace{-0.7cm}

\begin{table}[!t]
\centering
\small
{
\begin{tabular}{{c}|P{1.25cm}P{1.25cm}P{1.25cm}P{1.4cm}}
\hline
Model   & COSNet~\cite{cosnet} & F2Net~\cite{f2net} & MATNet~\cite{matnet} & AMCNet~\cite{amc}\\ 
\hline 
$\mathcal{J}$↑ & 75.6 &77.5 &76.1 &76.5\\ 
\hline 
\hline
Model   & IMP~\cite{imp} & PMN~\cite{pmn}  &  TMO~\cite{tmo}    & \textbf{Ours}\\ 
\hline 
$\mathcal{J}$↑  & 77.5 & 77.8  & \cellcolor{orange!25}\textbf{79.9}   & \cellcolor{yellow!25}79.1  \\ 
\hline
\end{tabular}
}
\vspace{-0.3cm}
\caption{Evaluation results on FBMS. \colorbox{orange!25}{\textbf{best}} and \colorbox{yellow!25}{second}.}
\label{tab:fbms}
\vspace{-0.8cm}
\end{table}

\subsection{Experimental Results}

\textbf{Evaluation on DAVIS 2016.} As shown in \tablename~\ref{tab:quantitative}, we evaluate our framework TSANet compared with state-of-the-art works. TSANet outperforms all other recent works. TSANet scores $87.4 \% \:\mathcal{J} \& \mathcal{F}\: \text {Mean}$. We do not use any post-processing techniques~\cite{crf}. We illustrate our qualitative results in \figurename~\ref{fig:qualitative}. 
TSANet maintains robust performance even in challenging scenes such as bmx-trees and kite-surf containing occluded and very thin objects. TSANet still shows the robustness and dense prediction results. For dance-twirl sequence which has many background distractors similar to the target object, our framework shows distinctive segmentation results although the motion of the target is dynamic.

\noindent
\textbf{Evaluation on FBMS.} The evaluation results for FBMS dataset are reported in \tablename~\ref{tab:fbms}. Following previous works~\cite{tmo,f2net,dtt}, we evaluate our performance on $\mathcal{J}$ score. We achieve $79.1 \% \:\text{over}\:\mathcal{J} \:\text {score}$ which is comparable with the state-of-the-art performance. TSANet also shows promising qualitative results as shown in \figurename~\ref{fig:qualitative}.

\noindent
\textbf{Ablation Study.}
As shown in \tablename~\ref{tab:ablation}, we demonstrate the effectiveness of each component of the proposed methods on DAVIS 2016. The efficiency of the backbone is important impact on model performance. TAF achieves 1.2\% improvements with ResNet50~\cite{resnet} and 0.9\% improvements with ResNet101 in $\mathcal{J} \& \mathcal{F} \: \text{Means}$. These improvements prove that TAF aims to utilize the useful information of adjacent frames focused on target object by aligning adjacent frames to the target frame regardless of the depth of the backbone. In addition, SAD also shows enhancement of the model performance. It demonstrates that SAD reduces the loss of information which is caused by deterministic algorithm,i.e.,bilinear upsampling. In EXP3, Seq means 2 step training which pre-train on DUTS and finetune on DAVIS 2016. It shows significant performance degradation. Our model achieves the best performance at image resolution of 512 $\times$ 512.  

\begin{table}[t!]
\small
{
\begin{tabular}{c|cccccc}
\hline
EXP & TAF &SAD & Backbone  & Res. & Training                & $\mathcal{J}$\&$\mathcal{F}$ \\ \hline  \hline
1   &  &        & R50  & 384        & Joint                   & 85.2 \\ 
2   &  &         & R101  & 384        & Joint                   & 86.1 \\ 
\hline
3   & \checkmark &     & R50  & 384        & Seq             & 85.1 \\ 
4   & \checkmark &     & R50  & 384        & Joint                   & 86.3 \\
5   & \checkmark &     & R101  & 384        & Joint                   & 87.0 \\
6   & \checkmark & \checkmark & R50  & 384        & Joint                   & 86.5 \\ 
\hline
7   & \checkmark &     & R101 & 512        & Joint                   & 87.3 \\ 
8   & \checkmark & \checkmark & R101 & 512        & Joint                   & 87.4 \\ 
\hline
\end{tabular}
}
\vspace{-0.1cm}
\caption{Ablation study of our methods on DAVIS 2016.}
\label{tab:ablation}
\vspace{-0.4cm}
\end{table}

\vspace{-0.4cm}
\section{Conclusion}
\label{sec:conclusion}
\vspace{-0.3cm}
In this paper, we propose TSANet, which consists of Temporal Alignment Fusion (TAF) and Scale Alignment Decoder (SAD). TAF aligns the information from adjacent frames to target frame for aggregating more useful information about target object. TAF reduces the dependency on a single optical flow corresponding to the target frame utilizing the information of adjacent frames. SAD aggregates feature maps for different scales and predicts the salient object mask more precisely in continuous manner. TSANet shows outstanding performance on the benchmark datasets, DAVIS 2016, and FBMS. We demonstrate that our network attains $87.4 \% \:\mathcal{J} \& \mathcal{F} \text { Mean }$ score on DAVIS 2016 with the state-of-the-art performance. In addition, we analyze the effectiveness of each components with ablation study.



\clearpage

\bibliographystyle{IEEEbib}
\footnotesize
\bibliography{strings,refs}

\end{document}